# Complexity Results and Approximation Strategies for MAP Explanations

**James D. Park**                                                    JD@CS.UCLA.EDU
**Adnan Darwiche**                                              DARWICHE@CS.UCLA.EDU
*Computer Science Department*
*University of California*
*Los Angeles, CA 90095*

## Abstract

MAP is the problem of finding a most probable instantiation of a set of variables given evidence. MAP has always been perceived to be significantly harder than the related problems of computing the probability of a variable instantiation (Pr), or the problem of computing the most probable explanation (MPE). This paper investigates the complexity of MAP in Bayesian networks. Specifically, we show that MAP is complete for $NP^{PP}$ and provide further negative complexity results for algorithms based on variable elimination. We also show that MAP remains hard even when MPE and Pr become easy. For example, we show that MAP is NP-complete when the networks are restricted to polytrees, and even then can not be effectively approximated.

Given the difficulty of computing MAP exactly, and the difficulty of approximating MAP while providing useful guarantees on the resulting approximation, we investigate best effort approximations. We introduce a generic MAP approximation framework. We provide two instantiations of the framework; one for networks which are amenable to exact inference (Pr), and one for networks for which even exact inference is too hard. This allows MAP approximation on networks that are too complex to even exactly solve the easier problems, Pr and MPE. Experimental results indicate that using these approximation algorithms provides much better solutions than standard techniques, and provide accurate MAP estimates in many cases.

## 1. Introduction

The task of computing the Maximum a Posteriori Hypothesis (MAP) is to find the most likely configuration of a set of variables given *partial* evidence about the complement of that set. The focus of this paper is on the complexity of computing MAP in Bayesian networks, and on a class of best effort methods for approximating MAP.

One specialization of MAP which has received a lot of attention is the Most Probable Explanation (MPE). MPE is the problem of finding the most likely configuration of a set of variables given *complete* evidence about the complement of that set. The primary reason for this attention is that MPE seems to be a much simpler problem than its MAP generalization. Unfortunately, MPE is not always suitable for the task of providing explanations. Consider for example the problem of system diagnosis, where each component has an associated variable representing its health. Given some evidence about the system behavior, one is usually interested in computing the most probable configuration of health variables. This is a MAP problem since the available evidence does not usually specify the value of each

non–health variable. It is common to approximate this problem using MPE, in which case one is finding the most likely configuration of every unknown variable, including health variables and some other variables of no particular interest, such as the inputs and outputs of system components. However, the projection of an MPE solution on health variables is usually not a most likely configuration. Neither is the configuration obtained by choosing the most likely state of each health variable separately.

MAP turns out to be a very difficult problem, even when compared to the MPE problem, or to the Pr problem for computing the probability of evidence. Specifically, we provide in Section 2 some complexity results which indicate that neither exact nor approximate solutions can be guaranteed for MAP, even under very restricted circumstances. Yet, MAP remains an important problem for which we would like to generate solutions. Therefore, we propose in Section 3 a general framework based on local search for best–effort approximation of MAP. We also provide two specific instances of the proposed framework, one is applicable to networks that are amenable to exact computation of Pr and is given in Section 3, while the other is applicable to networks that are not even amenable to Pr and is given in Section 4. We report on experimental results for each method using real–world and randomly generated Bayesian networks, which illustrate the effectiveness of our proposed framework on a wide range of networks. We close our paper by some concluding remarks in Section 5.

## 2. MAP Complexity

We begin this section by reviewing some complexity theory classes and terminology that pertain to the complexity of MAP. We then examine the complexity of MAP in the general case, followed by examining the complexity when the number of MAP variables is constrained. We then consider the complexity of MAP algorithms based on variable elimination. We conclude the complexity section by examining the complexity of MAP on polytrees.

### 2.1 Complexity Review

We assume that the reader is familiar with the basic notions of complexity theory like the hardness and completeness of languages, as well as the complexity class NP.

In addition to NP, we will also be interested in the class PP and a derivative of it. Informally, PP is the class which contains the languages for which there exists a nondeterministic Turing machine where the majority of the nondeterministic computations accept if and only if the string is in the language. PP can be thought of as the decision version of the functional class #P. As such, PP is a powerful language. In fact NP $\subseteq$ PP, and the inequality is strict unless the polynomial hierarchy collapses to the second level.[1]

Another idea we will need is the concept of an oracle. Sometimes it is useful to ask questions about what could be done if an operation were free. In complexity theory this is modeled as a Turing machine with an oracle. An oracle Turing machine is a Turing machine with the additional capability of being able to obtain answers to certain queries in a single time step. For example, we may want to designate the class of languages that could be recognized in nondeterministic polynomial time if any PP query could be answered for free. The class of languages would be NP with a PP oracle, which is denoted NP$^{\text{PP}}$.

---

1. This is a direct result of Toda's theorem (Toda, 1991). From Toda's theorem $P^{PP}$ contains the entire polynomial hierarchy (PH), so if NP = PP, then PH $\subseteq$ $P^{PP}$ = $P^{NP}$.

Consider now a Boolean expression $\phi$ over variables $X_1, \ldots, X_n$. The following three classical problems are complete for the above complexity classes:

**SAT:** Is there a truth assignment (world) that satisfies $\phi$? This problem is NP–complete.

**MAJSAT:** Do the majority of worlds satisfy $\phi$? This problem is PP–complete.

**E-MAJSAT:** Is there an instantiation of variables $X_1, \ldots, X_k$, $1 \leq k \leq n$, under which the majority of worlds satisfy $\phi$? This problem is NP$^{\text{PP}}$–complete.

Intuitively, to solve an NP–complete problem we have to *search* for a solution among an exponential number of candidates, where it is easy to decide whether a given candidate constitutes a solution. For example, in **SAT**, we are searching for a world that satisfies a sentence (testing whether a world satisfies a sentence can be done in time linear in the sentence size). To solve a PP–complete problem, we have to *add up* the weights of solutions, where it is easy to decide whether a particular candidate constitutes a solution and it is also easy to compute the weight of a solution. For example, in **MAJSAT**, a solution is a world that satisfies the sentence and the weight of a solution is 1. Finally, to solve an NP$^{\text{PP}}$–complete problem, we have to search for a solution among an exponential number of candidates, but we also need to solve a PP–complete problem in order to decide whether a particular candidate constitutes a solution. For example, in **E-MAJSAT**, we are searching for an instantiation $x_1, \ldots, x_k$, but to test whether an instantiation satisfies the condition we want, we must solve a **MAJSAT** problem.

## 2.2 Decision Problems

We will be dealing with the decision versions of Bayesian network problems in this paper, which we define formally in this section.

A Bayesian network is a pair $(G, \Theta)$, where $G$ is a directed acyclic graph (DAG) over variables $\mathbf{X}$, and $\Theta$ defines a conditional probability table (CPT) $\Theta_{X|\mathbf{U}}$ for each variable $X$ and its parents $\mathbf{U}$ in the DAG $G$. That is, for each value $x$ of variable $X$ and each instantiation $\mathbf{u}$ of parents $\mathbf{U}$, the CPT $\Theta_{X|\mathbf{U}}$ assigns a number in $[0, 1]$, denoted by $\theta_{x|\mathbf{u}}$, to represent the probability of $x$ given $\mathbf{u}$.[2] The probability distribution Pr induced by a Bayesian network $(G, \Theta)$ is given as follows. For a complete instantiation $\mathbf{x}$ of the network variables $\mathbf{X}$, the probability of $\mathbf{x}$ is given by

$$\Pr(\mathbf{x}) \stackrel{def}{=} \prod_{x\mathbf{u} \sim \mathbf{x}} \theta_{x|\mathbf{u}},$$

where $x\mathbf{u}$ is the instantiation of a family (a variable and its parents) and $\sim$ represents the compatibility relation among instantiations. That is, the probability assigned to a complete variable instantiation $\mathbf{x}$ is the product of all parameters that are consistent with that instantiation.

The following decision problems assume that we are given a Bayesian network $(G, \Theta)$ that has rational parameters and induces the probability distribution Pr. Moreover, by evidence $\mathbf{e}$, we mean an instantiation of variables $\mathbf{E}$.

---

2. Hence, we must have $\sum_x \theta_{x|\mathbf{u}} = 1$.

**D-MPE:** Given a rational number $p$, evidence $\mathbf{e}$, and the set of network variables $\mathbf{X}$, is there an instantiation $\mathbf{x}$ such that $\Pr(\mathbf{x}, \mathbf{e}) > p$?

**D-PR:** Given a rational number $p$ and evidence $\mathbf{e}$, is $\Pr(\mathbf{e}) > p$?

**D-MAP:** Given a rational number $p$, evidence $\mathbf{e}$, and some set of variables $\mathbf{Q}$, is there an instantiation $\mathbf{q}$ such that $\Pr(\mathbf{q}, \mathbf{e}) > p$? Variables $\mathbf{Q}$ are called the MAP variables in this case.

While decision problems are useful for examining the complexity of finding an exact solution, what we are really interested in is the functional problem of actually computing the solution. When we can't solve the problem exactly, we would also like to know how close we can get efficiently. For that we consider approximation algorithms. We now define the notion of an *approximation factor* which we will use when discussing the complexity of approximation algorithms. Specifically, we will say that an approximate solution $M'$ is within the approximation factor $\epsilon > 1$ of the true solution $M$ in case $\frac{M}{\epsilon} \leq M' \leq \epsilon M$. Moreover, we will say that an algorithm provides an $f(n)$–factor approximation in case for all problems of size $n$, the approximate solutions returned by the algorithm are within the approximation factor $f(n)$.

### 2.3 MAP Complexity for the General Case

Computing MPE, Pr, and MAP are all NP–Hard, but there still appears to be significant differences in their complexity. MPE is basically a combinatorial optimization problem. Computing the probability of a complete instantiation is trivial, so the only real difficulty is determining which instantiation to choose. **D-MPE** is NP-complete (Shimony, 1994). Pr is a completely different type of problem, characterized by counting instead of optimization, as we need to add up the probability of network instantiations. **D-PR** is PP-complete (Litmman, Majercik, & Pitassi, 2001)—notice that this is the complexity of the decision version, not the functional version which is #P-complete (Roth, 1996). MAP combines both the counting and optimization paradigms. In order to compute the probability of a particular instantiation, a Pr query is needed. Optimization is also required, in order to be able to decide between the many possible instantiations. This is reflected in the complexity of MAP.

**Theorem 1 D-MAP** *is* $\mathrm{NP}^{\mathrm{PP}}$-*complete.*[3]

*Proof:* Membership in $\mathrm{NP}^{\mathrm{PP}}$ is immediate. Given any instantiation $\mathbf{q}$ of the MAP variables, we can verify if it is a solution by querying the PP oracle if $\Pr(\mathbf{q}, \mathbf{e}) > k$.

To show hardness, we reduce **E-MAJSAT** (Littman, Goldsmith, & Mundhenk, 1998) to **D-MAP** by first creating a Bayesian network that models a Boolean formula $\phi$. For each variable $X_i$ in the formula $\phi$, we create an analogous variable in the network with values $\{T, F\}$ and a uniform prior probability. Then, for each logical operator, we create a variable with values $\{T, F\}$ whose parents are the variables corresponding to its operands, and whose CPT encodes the truth table for that operator (see Figure 1 for a simple example). Let $V_\phi$ be the network variable corresponding to the top level operand.

---

3. This result was stated without proof by Littman (1999).

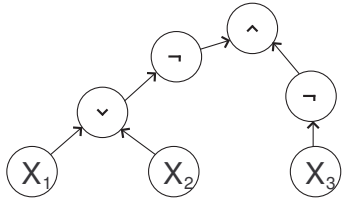

Figure 1: The Bayesian network produced using the reduction in Theorem 1 for Boolean formula $\neg(x_1 \lor x_2) \land \neg x_3$.

For a complete instantiation $\mathbf{x}$ of all of the variables $\mathbf{X}$ appearing in the Boolean expression $\phi$, with evidence $\mathbf{V}_\phi = T$, we have:

$$\Pr(\mathbf{x}, V_\phi = T) = \begin{cases} \frac{1}{2^n} & \mathbf{x} \text{ satisfies } \phi \\ 0 & \text{otherwise} \end{cases}$$

For a particular instantiation $\mathbf{q}$ of MAP variables $X_1, ..., X_k$, and evidence $V_\phi = T$, we have:

$$\begin{aligned} \Pr(\mathbf{q}, V_\phi = T) &= \sum_{x_{k+1},...,x_n} \Pr(\mathbf{q}, x_{k+1},..x_n) \\ &= \frac{\#\mathbf{q}}{2^n} \end{aligned}$$

where $\#\mathbf{q}$ is the number of complete variable instantiations compatible with $\mathbf{q}$ that satisfies $\phi$. Since there are $2^{n-k}$ possible instantiations of $X_{k+1}, ..., X_n$, the fraction $f_\mathbf{q}$ satisfied is $\#\mathbf{q}/2^{n-k}$, so

$$\Pr(\mathbf{q}, V_\phi = T) = \frac{f_\mathbf{q}}{2^k}$$

Thus, an instantiation $\mathbf{q}$ of the MAP variables is compatible with more than half of the complete, satisfying instantiations if $\Pr(\mathbf{q}, V_\phi = T) > 1/2^{k+1}$. So the MAP query over variables $X_1, ..., X_k$ with evidence $V_\phi = T$ and threshold $1/2^{k+1}$ is true if and only if the **E-MAJSAT** query is also true. $\square$

In fact, the above theorem can be strengthened.

**Theorem 2 D-MAP** *remains complete for* $\mathrm{NP}^{\mathrm{PP}}$ *even when (1) the network has depth 2, (2) there is no evidence, (3) all variables are Boolean, and (4) all network parameters lie in the interval* $[\frac{1}{2} - \epsilon, \frac{1}{2} + \epsilon]$ *for any fixed* $\epsilon > 0$.

The proof appears in Appendix A. Unlike computing probabilities, which becomes easy when the number of evidence nodes is bounded by a constant and the parameters are bounded away from 0 (it falls into RP as described by Dagum & Luby, 1997), MAP retains its $\mathrm{NP}^{\mathrm{PP}}$ complexity even under these restrictive circumstances.[4]

---

4. This is not altogether surprising since when evaluating the score of a possible solution, the MAP variables act as evidence variables.

NP$^{\text{PP}}$ is a powerful class, even compared to NP and PP. NP$^{\text{PP}}$ contains other important AI problems, such as probabilistic planning problems (Littman et al., 1998). The three classes are related by NP $\subseteq$ PP $\subseteq$ NP$^{\text{PP}}$, where the equalities are considered very unlikely. In fact, NP$^{\text{PP}}$ contains the entire polynomial hierarchy (Toda, 1991). Additionally, because MAP generalized Pr, MAP inherits the wild non–approximability of Pr. From the "Bayesian network simulates SAT" reduction we get:

**Corollary 3** *Approximating MAP within any approximation factor $f(n)$ is NP–hard.*

**Proof:** Using the evidence $V_\phi = T$, the exact MAP solution is the number of satisfying instantiations divided by $2^n$, which is 0 if it is unsatisfiable, and positive if it is satisfiable. If the formula is unsatisfiable, then the approximate solution must be 0 because $0/\epsilon = 0 \leq M' \leq \epsilon 0 = 0$, where $\epsilon = f(n)$. If the formula is satisfiable, then the approximate solution must be positive since $M/\epsilon > 0$. Thus we can test satisfiability by testing if the approximate MAP solution is zero or not.$\square$

## 2.4 Complexity Parameterized by the Number of Maximization Variables

We now examine the complexity when we restrict the number of maximization variables. Let $n$ be the number of non–evidence variables, and $k$ be the number of maximization variables. In the extreme case of $k = 0$, this is simply **D-PR**, so it is PP–complete. At the other extreme, when $k = n$, it becomes **D-MPE**, and is NP–complete. So constraining the number of maximization variables can have a dramatic impact on the complexity. We now examine this issue in detail. Let **D-MAP**$_m$ be the subset of **D-MAP** problems where $k = O(m)$, and let **D-MAP**$^m$ be the subset of **D-MAP** problems where $n - k = O(m)$. We can then consider the complexity of these parameterized classes of problems. The primary results are the following:

**Theorem 4** **D-MAP**$_{\log n}$ *is in* P$^{\text{PP}}$, *and* **D-MAP**$^{\log n}$ *is in* NP. *However, for any $\epsilon > 0$, both* **D-MAP**$_{n^\epsilon}$ *and* **D-MAP**$^{n^\epsilon}$ *remain* NP$^{\text{PP}}$*–complete.*

*Proof:* First, if $k = O(\log n)$, then the number of possible instantiations of the maximization variables is bounded by a polynomial. Thus, given a PP oracle, it is possible to decide the problem in polynomial time by asking for each instantiation $\mathbf{q}$ of the maximization variables whether $\Pr(\mathbf{q}, \mathbf{e})$ exceeds the threshold. Similarly, if $n - k = O(\log n)$, then for any instantiation $\mathbf{q}$ of the maximization variables, we can test to see if $\Pr(\mathbf{q}, \mathbf{e})$ exceeds the threshold by summing over the polynomial number of compatible instantiations.

For $k = O(n^\epsilon)$ we can provide a simple reduction to solve any **D-MAP** problem by creating a polynomially larger one satisfying the constraint on the number of maximization variables. From the unconstrained problem, we simply create a new problem by adding a polynomial number of irrelevant variables, with no parents or children. Similarly, we can provide a reduction of the general **D-MAP** problem to one constrained to have $n - k = O(n^\epsilon)$, by adding a polynomial number of maximization variables with no parents, no children, and deterministic priors. $\square$

### 2.5 Results for Elimination Algorithms

Solution to the general MAP problem seems out of reach, but what about for "easier" networks? State–of–the–art exact inference algorithms (variable elimination (Dechter, 1996), join trees (Lauritzen & Spiegelhalter, 1988; Shenoy & Shafer, 1986; Jensen, Lauritzen, & Olesen, 1990), recursive conditioning (Darwiche, 2001)) can compute $Pr$ and MPE in space and time complexity that is exponential only in the width of a given elimination order. This allows many networks to be solved using reasonable resources even though the general problems are very difficult. Similarly, state–of–the–art MAP algorithms can also solve MAP with time and space complexity that is exponential only in width of used elimination order but, for MAP, not all orders can be used. In this section, we investigate the complexity of variable elimination for MAP.

Before analyzing the complexity of variable elimination for MAP, we review variable elimination. First, we need the concept of a potential. A potential is simply a function over some subset of the variables, which maps each instantiation of its variables to a real number. The size of a potential is parameterized by the number of instantiations of its variables, and so is exponential in the number of variables. Notice that CPTs are potentials. In order to use variable elimination for Pr, MPE and MAP, we need three simple operations: multiplication, summing–out, and maximization. Multiplication of two potentials $\phi_1$ and $\phi_2$ with variables $\mathbf{XY}$ and $\mathbf{YZ}$ respectively (where $\mathbf{Y}$ is the set of variables they have in common), is defined as $(\phi_1\phi_2)(\mathbf{xyz}) = \phi_1(\mathbf{xy})\phi_2(\mathbf{yz})$. Notice that if both $\mathbf{X}$ and $\mathbf{Z}$ are nonempty, then the size of $\phi_1\phi_2$ is greater than the size of either $\phi_1$ or $\phi_2$. SumOut$(\phi, \mathbf{Y})$ where $\phi$ is over variables $\mathbf{XY}$ is defined as

$$\text{SumOut}(\phi, \mathbf{Y})(\mathbf{x}) = \sum_{\mathbf{y}} \phi(\mathbf{xy}),$$

where $\mathbf{y}$ ranges over the instantiations of $\mathbf{Y}$. Maximization is similar to summing out but maximizes out the unwanted variables:

$$\text{Maximize}(\phi, \mathbf{Y})(\mathbf{x}) = \max_{\mathbf{y}} \phi(\mathbf{xy}).$$

In order to handle evidence, we need the concept of an evidence indicator. The evidence indicator $\lambda_E$ associated with evidence $E = e$ is a potential over variable $E$ where $\lambda_E(e) = 1$, and is zero for all other values.

Given a variable ordering $\pi$, variable elimination can be used to compute the probability of evidence $\mathbf{e}$ as follows:

1. Initialize $P$ to contain the evidence indicators for $\mathbf{e}$ and all of the conditional probability tables.

2. For each variable $X$, according to order $\pi$,

   (a) Remove from $P$ all potentials mentioning $X$.

   (b) Let $M_X$ be the product of all of those potentials.

   (c) add SumOut$(M_X, X)$ to $P$.

3. Return the product of all potentials in $P$.

In each iteration, a variable $X$ is eliminated which leads to removing all mention of $X$ from $P$. By step 3, all variables have been removed, so the potentials remaining are just constants and the resulting product is a single number representing the probability of evidence **e**. MPE is computed in the same way, except the projection in step 2c is replaced by maximization. The complexity of variable elimination is linear in the number of variables and linear in the size of the largest potential $M_X$ produced in step 2b. The size of the largest potential varies significantly based on the elimination order. The *width* of an elimination order is simply $\log_2(\text{size}(M_X)) - 1$ where $M_X$ is the largest potential produced using that elimination order.[5] The *treewidth* of a Bayesian network is the minimum width of all elimination orders. For Pr and MPE, any elimination order can be used, so the complexity is linear in the number of variables and exponential in the treewidth. The same is not true for MAP. Variable elimination for MAP is similar to the other methods, but with an extra constraint. In step 2c, if $X$ is a MAP variable the projection is replaced with maximization. If it is not a MAP variable projection is used. The extra constraint is that not all orders are valid. Maximization and projection do not commute, and maximization must be performed last. This means that for an elimination order to be valid for performing MAP, when $X$ is a MAP variable, the potential in step 2b must not mention any non-MAP variables. In practice this is ensured by requiring that the elimination order eliminate all of the MAP variables last. This tends to produce elimination orders with widths much larger than those available for Pr and MPE, often placing exact MAP solutions out of reach.

In order to assess the magnitude of increase in width caused by restricting the elimination order, we randomly generated 100 Bayesian networks, each containing 100 variables, according to the first method in Appendix B. For each network, we computed the width using the min–fill heuristic (Kjaerulff, 1990; Huang & Darwiche, 1996). Then, we repeatedly added a single variable to the set of MAP variables, and computed the constrained width, again using min–fill, but eliminating the MAP variables last. This process was repeated until all variables were in the MAP variable set. Figure 2 contains statistics on these experiments. The $X$–axis corresponds to the number of MAP variables (thus $X = 0$ corresponds to the unconstrained width). The $Y$–axis corresponds to the width found. The graph details the minimum, maximum, mean, and weighted mean for each of the 100 networks. The weighted mean takes into account that the complexity is exponential in the width, and so provides a better representation of the average complexity. It was computed as $\log_2(\frac{1}{n}\sum_{i=1}^{n} 2^{w_i})$. Notice that the unconstrained widths range from 11 to 18, and that as the number of MAP variables increases, the width increases dramatically. For example, even when only a quarter of the variables are MAP variables ($X = 25$) the widths range between 22 and 34, (which corresponds roughly from difficult but doable to well beyond what today's inference algorithms can handle on today's computers) with a weighted average above 30. Notice also, that as we would expect from our complexity analysis, problems with very few or very many MAP variables are significantly easier than those in the middle range.

We now consider the question of whether there are less stringent conditions for valid elimination orders, that may allow for orders with smaller widths.

---

5. The -1 in the definition is to preserve compatibility with the previously defined notion of treewidth in graph theory.

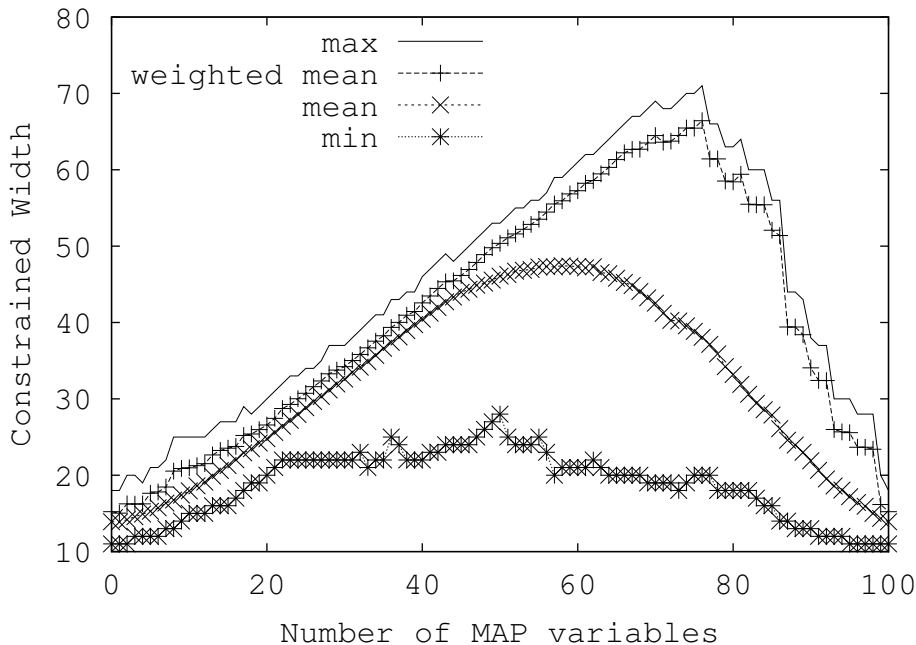

Figure 2: Statistics on the constrained width for different numbers of MAP variables. The $X$–axis is the number of MAP variables, and the $Y$–axis is the width. Notice that the widths required for the general MAP problem are significantly larger than for Pr or MPE, which correspond to $X = 0$ and $X = 100$ respectively.

As described earlier, given an ordering, elimination algorithms work by stepping through the ordering, collecting the potentials mentioning the current variable, multiplying them, then replacing them with the potential formed by summing out (or maximizing) the current variable from the product. This process can be thought to induce an *evaluation tree;* see Figure 3. The evaluation tree for an elimination order $\pi$ is described as follows. The leaves correspond to the CPTs of given Bayesian network, and the internal nodes correspond to potentials created during the elimination process. The children of a potential $\phi$ represent other potentials that had to be multiplied together when constructing $\phi$. Note that each internal node in the elimination tree corresponds to variable in the order $\pi$, whose elimination leads to constructing the node; Figure 3(b). Therefore, an evaluation tree can be viewed as inducing a partial elimination order; see Figure 3(c).

The standard way of constructing a valid elimination order for MAP is to eliminate the MAP variables $\mathbf{Q}$ last. Two questions present themselves. First, are there valid orderings in which variables $\mathbf{Q}$ are not eliminated last? And second, if so, can they produce widths smaller than those generated by eliminating variables $\mathbf{Q}$ last?

The answer to the first question is yes, there are other valid elimination orders in which variables $\mathbf{Q}$ are not eliminated last. To see that, suppose we have a variable order $\pi$ which induces a particular evaluation tree $T$, and let $\sigma$ be the partial elimination order corresponding to $T$. Any variable order $\pi'$ which is consistent with the partial order $\sigma$ will also induce the tree $T$. Hence, if order $\pi$ was valid, then order $\pi'$ will also be valid. Figure 3 shows the evaluation tree induced by using the order $A, E, B, D, C$ for computing MAP over variables $\mathbf{Q} = C, D$. The order $A, B, D, E, C$ is consistent with this evaluation tree and, hence, is also valid for computing MAP over variables $C, D$. Yet, variables $C, D$ do not appear last in the order.

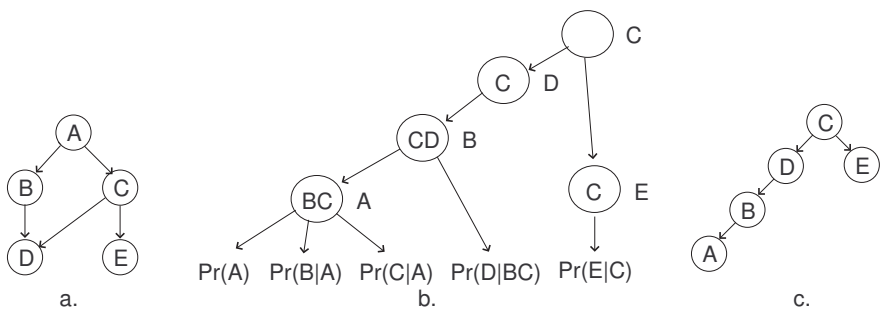

Figure 3: (a) A Bayesian network, (b) an evaluation tree for the elimination order $A, E, B, D, E$, and (c) a partial elimination order induced by the evaluation tree.

Unfortunately, orders in which variables $\mathbf{Q}$ are not eliminated last do not help.

**Theorem 5** *For any elimination order $\pi$ which is valid for computing MAP over variables $\mathbf{Q}$, there is an ordering of the same width in which variables $\mathbf{Q}$ are eliminated last.*

*Proof:* Consider the evaluation tree induced by any valid elimination order, and the corresponding partial order it induces. No variable in $\overline{\mathbf{Q}}$ is the parent of any variable in $\mathbf{Q}$. To prove this, suppose that $X$ is a parent of $Y$ in the evaluation tree, where $\mathbf{X} \in \overline{\mathbf{Q}}$ and $\mathbf{Y} \in \mathbf{Q}$. This means that the potential which results from eliminating variable $Y$ includes variable $X$, which also means that $X$ must have appeared in some of the potentials we multiplied when elimination variable $Y$. But this is a contradiction since the evaluation tree and its underlying order are valid. Since no variable in $\overline{\mathbf{Q}}$ is a parent of a variable in $\mathbf{Q}$, all variables in $\overline{\mathbf{Q}}$ can be eliminated first in any order consistent with the partial order defined by the evaluation tree. Then, all variables in $\mathbf{Q}$ can be eliminated, again obeying the partial ordering defined by the evaluation tree. Because the produced order has the same elimination tree as the original order, they have the same width. $\square$

## 2.6 MAP on Polytrees

Theorem 5 has significant complexity implications for elimination algorithms even on polytrees.

**Theorem 6** *Elimination algorithms require exponential resources to perform MAP, even on some polytrees.*

*Proof:* Consider computing MAP over variables $X_1, \ldots, X_n$ given evidence $S_n = T$ for the network shown in Figure 4. By Theorem 5, there is no order whose width is smaller than that of an order $\pi$ in which we eliminate variables $S_0, \ldots, S_n$ first, and then variables $X_1, \ldots, X_n$ last. It is easy to show though that any such order $\pi$ will have width $n$. Hence, variable elimination will require exponential resources using such an order. □

The set of MAP variables makes a crucial difference in the complexity of MAP computations. For example, if the MAP variables were $X_1, \ldots, X_{n/2}, S_0, \ldots, S_{n/2}$ instead of $X_1, \ldots, X_n$ can be solved in linear time.

The above negative findings are specific to variable elimination algorithms. The question then is whether this difficulty is an idiosyncrasy of variable elimination which can be avoided if we were to use some other method for computing MAP. The following result, however, shows that finding a good general algorithm for MAP on polytrees is unlikely.

**Theorem 7** *MAP is NP–complete when restricted to polytrees.*

*Proof:* Membership is immediate. Given a purported solution instantiation $\mathbf{q}$, we can compute $\Pr(\mathbf{q}, \mathbf{e})$ in linear time and test it against the bound. To show hardness, we reduce **MAXSAT** to MAP on a polytree.[6] Similar reductions were used by Papadimitriou and Tsitsiklis (1987) and Littman et al. (1998) relating to partially observable Markov decision problems, and probabilistic planning respectively. The **MAXSAT** problem is defined as follows:

> Given a set of clauses $C_1, ..., C_m$ over variables $Y_1, ..., Y_n$ and an integer bound $k$, is there an assignment of the variables, such that more than $k$ clauses are satisfied.

The idea behind the reduction is to model the random selection of a clause, and then successively checking whether the instantiation of each variable satisfies the selected clause. The clause selector variable $S_0$ with possible values $1, 2, ..., m$ has a uniform prior. Each propositional variable $Y_i$ induces two network variables $X_i$ and $S_i$, where $X_i$ represents the value of $Y_i$, and has a uniform prior, and $S_i$ represents whether any of $Y_1, ..., Y_i$ satisfy the selected clause. $S_i = 0$ indicates that the selected clause was satisfied by one of $Y_1, ..., Y_i$. $S_i = c > 0$ indicates that the selected clause $C_c$ was not satisfied by $Y_1, ..., Y_i$. The parents of $S_i$ are $X_i$ and $S_{i-1}$ (the topology is shown in Figure 4). The CPT for $S_i$, for $i \geq 1$ is defined as

$$Pr(s_i | x_i, s_{i-1}) = \begin{cases} 1 & \text{if } s_i = s_{i-1} = 0 \\ 1 & \text{if } s_i = 0 \text{ and } s_{i-1} = j, \text{ and} \\ & \quad x_i \text{ satisfies } c_j \\ 1 & \text{if } s_i = s_{i-1} = j \text{ and } x_i \text{ does} \\ & \quad \text{not satisfy } c_j \\ 0 & \text{otherwise} \end{cases}$$

In words, if the selected clause was not satisfied by the first $i - 1$ variables ($s_{i-1} \neq 0$), and $x_i$ satisfies it, then $S_i$ becomes satisfied ($s_i = 0$). Otherwise, we have $s_i = s_{i-1}$. Now, for a particular instantiation $s_0$ of $S_0$, and instantiation $\mathbf{x}$ of variables $X_1, ..., X_n$,

$$\Pr(s_0, \mathbf{x}, S_n = 0) = \begin{cases} 1/(m2^n) & \text{if } \mathbf{x} \text{ satisfies clause } C_{s_0}; \\ 0 & \text{otherwise.} \end{cases}$$

---

6. Actually, we only need to reduce it to SAT, but the **MAXSAT** result will be used in Theorem 8.

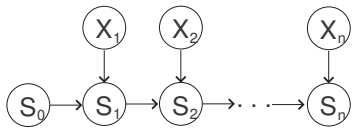

Figure 4: The network used in the reduction of Theorem 7.

Summing over $S_0$ yields $Pr(\mathbf{x}, S_n = 0) = \#_C/(m2^n)$ where $\#_C$ is the number of clauses that $\mathbf{x}$ satisfies. Thus MAP over $X_1, ..., X_n$ with evidence $S_n = 0$ and bound $k/(m2^n)$ solves the **MAXSAT** problem as well. □

Since **MAXSAT** has no polynomial time approximation scheme (unless P = NP), no polynomial time approximation scheme exists for MAP on polytrees. In fact, approximating MAP on polytrees appears to be much harder than approximating MAXSAT.

**Theorem 8** *Approximating MAP on polytrees within a factor of $2^{n^\epsilon}$ is NP-hard for any fixed $\epsilon$, $0 \le \epsilon < 1$, where $n$ is the size of the problem.*

The proof appears in Appendix A. So, not only is it hard to approximate within a constant factor, it is hard to approximate within a polynomial factor, or even a subexponential factor.

We close this section by a summary of the complexity results in this section:

- MAP is $\mathrm{NP}^{\mathrm{PP}}$–complete for arbitrary Bayesian networks, even if we have no evidence, every variable is binary, and the parameters are arbitrarily close to $1/2$.

- It is NP–hard to approximate MAP within any factor $f(n)$.

- Variable elimination for MAP requires exponential time, even on some polytrees.

- MAP is NP–complete for networks with polytree structure.

- Approximating MAP on polytrees within a factor of $2^{n^\epsilon}$ is NP–hard for any fixed $\epsilon \in [0, 1)$.

## 3. Approximating MAP when Inference is Easy

Since exact MAP computation is often intractable, approximation techniques are needed. Unfortunately, in spite of MAP's utility, relatively little work has been done in approximating it. In fact, there are only two previous methods for approximating MAP which we are aware of. The first (Dechter & Rish, 1998) uses the mini–bucket technique. The other (de Campos, Gamez, & Moral, 1999) uses genetic algorithms to approximate the best k configurations of the MAP variables (this problem is known as partial abduction). Practitioners typically resort to one of two simple approximation methods. One common approximation technique is to compute an MPE instantiation and then project the result on the MAP variables. That is, if we want to compute MAP for variables $\mathbf{S}$ given evidence $\mathbf{e}$, and if $\mathbf{S}'$ is the complement of variables $\mathbf{S} \cup \mathbf{E}$, we compute an instantiation $\mathbf{s}, \mathbf{s}'$ that maximizes $\Pr(\mathbf{s}, \mathbf{s}' \mid \mathbf{e})$ and then return $\mathbf{s}$. The other method computes posterior marginals for the MAP variables, $\Pr(S \mid \mathbf{e}), S \in \mathbf{S}$, and then choose the most likely state $s$ of each variable given $\mathbf{e}$.

We propose a general framework for approximating MAP. MAP consists of two problems that are hard in general—optimization and inference. A MAP approximation algorithm can be produced by substituting approximate versions of either the optimization or inference component (or both). The optimization problem is defined over the MAP variables, and the score for each solution candidate instantiation $\mathbf{s}$ of the MAP variables is the (possibly approximate) probability $\Pr(\mathbf{s}, \mathbf{e})$ produced by the inference method. This allows solutions tailored to the specific problem. For networks whose treewidth is manageable, but contains a hard optimization component (e.g. the polytree examples discussed previously), exact structural inference can be used, coupled with an approximate optimization algorithm. Alternatively, if the optimization problem is easy (e.g. there are few MAP variables) but the network isn't amenable to exact inference, an exact optimization method could be coupled with an approximate inference routine. If both components are hard, both the optimization and inference components need to be approximated.

We investigate in this section a family of approximation algorithms based on local search. We first consider the case when inference is tractable, then develop an extension to handle the case when inference is infeasible. The local search algorithms work basically as follows:

1. Start from an initial guess $\mathbf{s}$ at the solution.

2. Iteratively try to improve the solution by moving to a better neighbor $\mathbf{s}'$: $\Pr(\mathbf{s}' \mid \mathbf{e}) > \Pr(\mathbf{s} \mid \mathbf{e})$, or equivalently $\Pr(\mathbf{s}', \mathbf{e}) > \Pr(\mathbf{s}, \mathbf{e})$.

A *neighbor* of instantiation $\mathbf{s}$ is defined as an instantiation which results from changing the value of a single variable $X$ in $\mathbf{s}$. If the new value of $X$ is $x$, we will denote the resulting neighbor by $\mathbf{s} - X, x$. In order to perform local search efficiently, we need to compute the scores for all of the neighbors $\mathbf{s} - X, x$ efficiently. That is, we need to compute $\Pr(\mathbf{s} - X, x, \mathbf{e})$ for each $X \in \mathbf{S}$ and each of its values $x$ not in $\mathbf{s}$. If variables have binary values, we will have $\mid \mathbf{S} \mid$ neighbors in this case.

Local search has been proposed as a method for approximating MPE (Kask & Dechter, 1999; Mengshoel, Roth, & Wilkins, 2000). For MPE, the MAP variables $\mathbf{S}$ contain all variables which are not in $\mathbf{E}$ (the evidence variables). Therefore, the score of a neighbor, $\Pr(\mathbf{s} - X, x, \mathbf{e})$, can be computed easily since $\mathbf{s} - X, x, \mathbf{e}$ is a complete instantiation. In fact, given that we have computed $\Pr(\mathbf{s}, \mathbf{e})$, the score $\Pr(\mathbf{s} - X, x, \mathbf{e})$ can be computed in constant time.[7]

Unlike MPE, computing the score of a neighbor, $\Pr(\mathbf{s} - X, x, \mathbf{e})$, in MAP requires a global computation since $\mathbf{s} - X, x, \mathbf{e}$ may not be a complete instantiation. One of the main observations underlying our approach, however, is that the score $\Pr(\mathbf{s} - X, x, \mathbf{e})$ can be computed in $O(n \exp(w))$ time and space, where $n$ is the number of network variables and $w$ is the width of an arbitrary elimination order, i.e., we can use any elimination order for this purpose, no need for any constraints. In fact, we can even do better than this by computing the scores of all neighbors, $\Pr(\mathbf{s} - X, x, \mathbf{e})$ for all $X \in \mathbf{S}$ and every value $x$ of $X$, in $O(n \exp(w))$ time and space. Thus, if we have an elimination order of width $w$ for the given Bayesian network, then we can perform each search step in $O(n \exp(w))$ time and space. As we shall see later, it takes a small number of search steps to obtain a good MAP solution. Hence, the overall runtime is often $O(n \exp(w))$ too. Therefore, we can produce good quality MAP approximations in time and space which are exponential in the unconstrained width instead of the constrained one, which is typically much larger.

---

7. This assumes that none of the entries in the CPTs are 0. If there are 0 entries in the CPTs, it may take time linear in the number of network variables to compute the score. $\Pr(\mathbf{s}, \mathbf{e})$ is the product of the single entry of each CPT that is compatible with $\mathbf{s}, \mathbf{e}$. When changing the state of variable $X$ from $x$ to $x'$, the only values in the product that change are those from the CPTs of $X$ and its children. If none of the CPT entries are 0, $\Pr(\mathbf{s} - X, x', \mathbf{e})$ can be computed by dividing $\Pr(\mathbf{s}, \mathbf{e})$ by the old and multiplying by the new entry for the CPTs for $X$ and its children. This can be done in constant time if the number of children is bounded by a constant.

The local search method proposed in this section differs from the local search methods used for MPE in that the unconstrained width must be small enough so that a search step can be performed relatively efficiently. It is pointless to use this method to approximate MPE since in the time to take one step, the MPE could be computed exactly. *This method is applicable when the unconstrained width is reasonable but the constrained width is not (see Figure 2).*

## 3.1 Computing Neighbor Scores Efficiently

The key to computing the neighbor scores efficiently is to express the inference problem as a function over the evidence indicators. For each state $x$ of variable $X$, the evidence indicator $\lambda_x$ is one if it is compatible with the evidence, and zero otherwise. This is a common technique that is typically used to allow the modeling of a wider range of evidence assertions. For example, this allows evidence assertions such as $X \neq x$ by setting $\lambda_x = 0$, and the remaining indicators for $X$ to one. We will use them for a different purpose however. When the inference problem is cast as a function $f$ of the evidence indicators ($f(\lambda_{\mathbf{e}}) = \Pr(\mathbf{e})$, where $\lambda_{\mathbf{e}}$ consists of all of the evidence indicators, set so that they are compatible with $\mathbf{e}$), then $\frac{\partial f}{\partial \lambda_x}(\lambda_{\mathbf{e}}) = \Pr(\mathbf{e} - X, x)$. When our we add the current state to the evidence, this partial derivative yields $\Pr(\mathbf{s} - X, x, \mathbf{e})$, which is precisely the score for one of the neighbors.

We can use the jointree algorithm (Park & Darwiche, 2003), or the differential inference approach (Darwiche, 2003) to compute all of the partial derivatives efficiently. In the differential approach, these values are immediate, as the entire approach is based on evaluating and differentiating the expression $f$ above. It can also be computed using jointrees by using the Shenoy–Shafer propagation scheme. Specifically, an evidence indicator table is added for each variable, and evidence about that variable is entered by setting the appropriate indicator entries. The partial derivatives of all of the indicators associated with a variable are obtained by multiplying all other tables assigned to the same cluster, and all messages into the cluster, then projecting that product onto the variable. In either case, the partial derivatives for all indicators, and thus the score for all neighbors, can be computed in $O(n \exp(w))$ time, which is the same complexity as simply computing the score for the current state.

## 3.2 Search Methods

We tested two common local search methods, *stochastic hill climbing* and *taboo search*. Stochastic hill climbing proceeds by repeatedly either changing the the state of the variable that creates the maximum probability change, or changing a variable at random. Figure 5 gives the algorithm explicitly.

Figure 5: Stochastic hill climbing algorithm.

Taboo search is similar to hill climbing except that the next state is chosen as the best state that hasn't been visited recently. Because the number of iterations is relatively small we save all of the previous states so that at each iteration a unique point is chosen. Pseudocode for taboo search appears in Figure 6.

## 3.3 Initialization

The quality of the solution returned by a local search routine depends to a large extent on which part of the search space it is given to explore. We implemented several algorithms to compare the solution quality with different initialization schemes. Suppose that $n$ is the number of network variables, $w$ is the width of a given elimination order, and $m$ is the number of MAP variables.

1. *Random initialization (Rand).* For each MAP variable, we select a value uniformly from its set of states. This method takes $O(m)$ time.

2. *MPE based initialization (MPE).* We compute the MPE solution given the evidence. Then, for each MAP variable, we set its value to the value that the variable takes on in the MPE solution. This method takes $O(n \exp(w))$ time.

3. *Maximum likelihood initialization (ML).* For each MAP variable $X$, we set its value to the instance $x$ that maximizes $Pr(x \mid \mathbf{e})$. This method takes $O(n \exp(w))$ time.

4. *Sequential initialization (Seq).* This method considers the MAP variables $X_1, \ldots, X_m$, choosing each time a variable $X_i$ that has the highest probability $\Pr(x_i \mid \mathbf{e}, \mathbf{y})$ for one

---

<u>Given:</u> Probability distribution Pr, evidence $\mathbf{e}$, MAP variables $\mathbf{S}$.
<u>Compute:</u> An instantiation $\mathbf{s}$ which (approximately) maximizes $Pr(\mathbf{s} \mid \mathbf{e})$.

Initialize current state $\mathbf{s}$.
$\mathbf{s}_{best} = \mathbf{s}$
Repeat many times
    Add $\mathbf{s}$ to *visited*
    Compute the score $Pr(\mathbf{s} - X, x, \mathbf{e})$ for each neighbor $\mathbf{s} - X, x$.
    $\mathbf{s} = \mathbf{s}'$ where $\mathbf{s}'$ is a neighbor with the highest score not in *visited*.
    If no such neighbor exists (this rarely occurs)
        Repeat for several times
            $\mathbf{s} = \mathbf{s}'$ where $\mathbf{s}'$ is a randomly selected neighbor of $\mathbf{s}$.
    If $Pr(\mathbf{s}, \mathbf{e}) > Pr(\mathbf{s}_{best}, \mathbf{e})$ then
        $\mathbf{s}_{best} = \mathbf{s}$
Return $\mathbf{s}_{best}$

---

Figure 6: Taboo search. Notice that the action taken is to choose the best neighbor that hasn't been visited. This leads to moves that decrease the score after a peak is discovered.

of its values $x_i$, where $\mathbf{y}$ is the instantiation of MAP variables considered so far. This method takes $O(mn \exp(w))$ time.

## 3.4 Experimental Results

Two search methods (*Hill* and *Taboo*) and four initialization methods (*Rand*, *MPE*, *ML*, *Seq*) lead to 8 possible algorithms. Each of the initialization methods can also be viewed as an approximation algorithm since one can simply return the computed initialization. This leads to a total of 12 different algorithms. We experimentally evaluated and compared 11 of these algorithms, leaving out the algorithm corresponding to random initialization.

We tested the algorithms on various synthetically generated data sets as well as real world networks. For the synthetic networks, we generated random network structures using two generation methods (see Appendix B). For each structure, we quantified the CPTs for different bias coefficients from 0 (deterministic except the roots), to .5 (values chosen uniformly) so we could evaluate the influence of CPT quantification on the solution quality. Each network consisted of 100 variables, with some of the root variables chosen as the MAP variables. If there were more than 25 root variables, we randomly selected 25 of them for the MAP variables. Otherwise we used all of the root variables. We chose root nodes for MAP variables because typically some subset of the root nodes are the variables of interest in diagnostic applications. Evidence was set by instantiating leaf nodes. Care was taken to insure that the instantiation had a non zero probability. Each algorithm was allowed 150 network evaluations.[8] We computed the true MAP and compared it to the solutions found by each algorithm. Additionally, we measured the number of network evaluations needed to find the solution each algorithm subsequently returned, and the number of peaks discovered

---

8. An evaluation takes $O(n \exp(w))$ time and space, where $n$ is the number of network variables and $w$ is the width of given elimination order.

Data Set 1 Solution Quality

|          | 0    | .125 | .250 | .375 | .5   |
|---------:|------|------|------|------|------|
| Rand-Hill | 147 | 805 | 917 | 946 | 966 |
| Rand-Taboo | 181 | 969 | 985 | 993 | 995 |
| ML | 526 | 497 | 676 | 766 | 817 |
| ML-Hill | 920 | 947 | 989 | 993 | 997 |
| ML-Taboo | 942 | 988 | 999 | 999 | **1000** |
| MPE | 999 | 333 | 160 | 127 | 100 |
| MPE-Hill | 999 | 875 | 923 | 952 | 973 |
| MPE-Taboo | **1000** | 986 | 992 | 990 | 998 |
| Seq | 930 | 965 | 990 | 999 | 997 |
| Seq-Hill | 941 | 971 | 992 | 999 | 997 |
| Seq-Taboo | 962 | **998** | **1000** | **1000** | **1000** |

Table 1: The solution quality of each method for the first data set. The number associated with each method and bias is the number of instances solved correctly out of 1000. The best scores for each bias are shown in bold.

Data Set 2 Solution Quality

|          | 0    | .125 | .250 | .375 | .5   |
|---------:|------|------|------|------|------|
| Rand-Hill | 20 | 634 | 713 | 799 | 845 |
| Rand-Taboo | 20 | 851 | 907 | 943 | 965 |
| ML | 749 | 453 | 495 | 519 | 514 |
| ML-Hill | 966 | 922 | 947 | 963 | 962 |
| ML-Taboo | 973 | 960 | 986 | 987 | 990 |
| MPE | 858 | 505 | 365 | 275 | 206 |
| MPE-Hill | 961 | 853 | 850 | 874 | 891 |
| MPE-Taboo | 978 | 952 | 962 | 977 | 980 |
| Seq | 988 | 955 | 964 | 985 | 972 |
| Seq-Hill | 988 | 960 | 966 | 986 | 976 |
| Seq-Taboo | **994** | **977** | **990** | **994** | **994** |

Table 2: The solution quality of each method for the second data set. The number associated with each method and bias is the number of instances solved correctly out of 1000. The best scores for each bias are shown in bold.

before that solution was discovered. The hill climbing method used in these data sets is pure hill climbing with random walk restart. That is, it hill climbs until it reaches a peak, then randomly flips some of the values to move to a new location.

We generated 1000 random network structures for each of the two structural generation methods. For each random structure generated, and each quantification method, we quantified the network, computed the exact MAP, and applied each of the approximation algorithms. Tables 1 and 2 show the solution quality of each of the methods by reporting the fraction of networks that were solved correctly; that is, the approximate answer had the same value as the exact answer.

One can draw a number of observations based on these experiments:

- In each case, taboo search performed slightly better than hill climbing with random restarts.

- The search methods were typically able to perform much better than the initialization alone.

- Even from a random start, the search methods were able to find the optimal solution in the majority of the cases.

- Overall, taboo search with sequential initialization performed the best, but required the most network evaluations.

Table 3 contains some statistics on the number of network evaluations (including those used for initialization) needed to achieve the value that the method finally returned. The mean number of evaluations is quite small for all of the methods. Surprisingly, for the hill climbing methods, the maximum is also quite small. In fact, after analyzing the results we discovered that the hill climbing methods never improved over the first peak they discovered.[9] This suggests that one viable method for quick approximation is to simply climb to the first peak and return the result. Taboo search on the other hand was able to improve on the first peak in some cases.

We ran ten MAP queries for each real world network we tested. For each query we randomly selected one fourth of the nodes to be the variables of interest, and selected one fourth of the nodes to be evidence nodes. The evidence values were chosen uniformly among the nonzero configurations. As our previous experiments demonstrated that a large number of iterations rarely helps, we reduced the number of iterations to 30. Also, we moved away from hill climbing with random restart to stochastic hill climbing (performing a random move with probability .35) since in our previous experiments the random restart never helped. Also, we ran a mini–bucket approximation algorithm (the only other MAP approximation algorithm we are aware of that is not subsumed by our technique) to compare its performance to our algorithms. Since exact MAP computations on these networks is too hard for current algorithms to handle, we compare the algorithms based on relative performance only.

Table 4 shows the number of times (out of ten) that each algorithm was able to produce the highest probability configuration discovered. The search based methods again performed much better than the other algorithms. Note that each of them outperformed the mini–bucket approximations on each network. Table 5 provides more specific details about the relative performance for each network. Each block contains the count of the number of times that each method produced solutions within some range of the best found solution.

---

9. It appears that the random walk used in restarting does not make eventually selecting a better region very likely when using so few search steps. Often, when a sub optimal hill was encountered, the optimal hill was just 2 or 3 moves away. In those cases, the taboo search was usually able to find it (because its search was more guided), while random walking was not.

Evaluations Required

| Method | Mean | Stdev | Max |
|---|---|---|---|
| Rand Hill | 12.5 | 2.5 | 21 |
| Rand Taboo | 14.3 | 11.0 | 144 |
| MPE | 1 | 0 | 1 |
| MPE Hill | 2.6 | 1.3 | 8 |
| MPE Taboo | 4.0 | 8.3 | 137 |
| ML | 1 | 0 | 1 |
| ML Hill | 1.6 | .74 | 4 |
| ML Taboo | 1.9 | 3.3 | 62 |
| Seq | 25 | 0 | 25 |
| Seq Hill | 25.0 | .04 | 26 |
| Seq Taboo | 25.0 | .9 | 45 |

Table 3: Statistics on the number of evaluations each method required before achieving the value it eventually returned. These are based on the random method 2, bias .5 data set. The statistics for the other data sets are similar.

| | MPE | | | ML | | | Seq | | | MB | | |
|---|---|---|---|---|---|---|---|---|---|---|---|---|
| | I | H | T | I | H | T | I | H | T | 14 | 16 | 18 |
| Barley | 3 | 9 | 8 | 3 | 10 | 9 | 7 | 10 | 10 | 1 | 3 | 5 |
| Mildew | 6 | 10 | 10 | 8 | 10 | 10 | 8 | 10 | 10 | 4 | 4 | 7 |
| Munin2 | 6 | 10 | 10 | 10 | 10 | 10 | 10 | 10 | 10 | 4 | 5 | 7 |
| Munin3 | 9 | 10 | 10 | 10 | 10 | 10 | 10 | 10 | 10 | 4 | 6 | 2 |
| Pigs | 0 | 0 | 0 | 5 | 9 | 9 | 8 | 8 | 8 | 3 | 1 | 6 |
| Water | 9 | 10 | 10 | 8 | 10 | 10 | 10 | 10 | 10 | 6 | 6 | 9 |

Table 4: Number of times out of ten that each algorithm found the instantiation that yielded the highest score. I, H, and T refer to initialization only, hill climbing, and taboo search respectively.

So for example, in the Barley group, in the MPE row, for the column labeled "> .5" there is a 3, indicating that in 3 of the 10 cases the solution found was between .5 and .9 times the best solution found for that query.

Qualitatively, these results are very similar to those obtained for the random networks. Again the search methods outperformed the static initialization methods. Note that for different networks, different initializations perform better. Notice also, that the search methods significantly outperformed the mini–bucket approximations in every network.

119

**Barley network results**

| Method | Best | > .9 | > .5 | > .01 | ≤ .01 |
|---|---|---|---|---|---|
| MPE | 3 | 2 | 3 | 2 | 0 |
| Hill | 9 | 0 | 0 | 1 | 0 |
| Taboo | 8 | 0 | 1 | 1 | 0 |
| ML | 3 | 2 | 2 | 0 | 3 |
| Hill | 10 | 0 | 0 | 0 | 0 |
| Taboo | 9 | 0 | 1 | 0 | 0 |
| Seq | 7 | 3 | 0 | 0 | 0 |
| Hill | 10 | 0 | 0 | 0 | 0 |
| Taboo | 10 | 0 | 0 | 0 | 0 |
| MB 14 | 1 | 2 | 1 | 3 | 3 |
| MB 16 | 3 | 3 | 1 | 2 | 1 |
| MB 18 | 5 | 2 | 0 | 3 | 0 |

**Mildew network results**

| Method | Best | > .9 | > .5 | > .01 | ≤ .01 |
|---|---|---|---|---|---|
| MPE | 6 | 1 | 3 | 0 | 0 |
| Hill | 10 | 0 | 0 | 0 | 0 |
| Taboo | 10 | 0 | 0 | 0 | 0 |
| ML | 8 | 1 | 1 | 0 | 0 |
| Hill | 10 | 0 | 0 | 0 | 0 |
| Taboo | 10 | 0 | 0 | 0 | 0 |
| Seq | 9 | 1 | 0 | 0 | 0 |
| Hill | 10 | 0 | 0 | 0 | 0 |
| Taboo | 10 | 0 | 0 | 0 | 0 |
| MB 14 | 4 | 1 | 0 | 1 | 4 |
| MB 16 | 4 | 0 | 0 | 1 | 5 |
| MB 18 | 7 | 0 | 1 | 0 | 2 |

**Munin2 network results**

| Method | Best | > .9 | > .5 | > .01 | ≤ .01 |
|---|---|---|---|---|---|
| MPE | 6 | 0 | 4 | 0 | 0 |
| Hill | 10 | 0 | 0 | 0 | 0 |
| Taboo | 10 | 0 | 0 | 0 | 0 |
| ML | 10 | 0 | 0 | 0 | 0 |
| Hill | 10 | 0 | 0 | 0 | 0 |
| Taboo | 10 | 0 | 0 | 0 | 0 |
| Seq | 10 | 0 | 0 | 0 | 0 |
| Hill | 10 | 0 | 0 | 0 | 0 |
| Taboo | 10 | 0 | 0 | 0 | 0 |
| MB 14 | 4 | 0 | 1 | 2 | 3 |
| MB 16 | 5 | 0 | 1 | 2 | 2 |
| MB 18 | 7 | 0 | 0 | 1 | 2 |

**Munin3 network results**

| Method | Best | > .9 | > .5 | > .01 | ≤ .01 |
|---|---|---|---|---|---|
| MPE | 9 | 0 | 0 | 1 | 0 |
| Hill | 10 | 0 | 0 | 0 | 0 |
| Taboo | 10 | 0 | 0 | 0 | 0 |
| ML | 10 | 0 | 0 | 0 | 0 |
| Hill | 10 | 0 | 0 | 0 | 0 |
| Taboo | 10 | 0 | 0 | 0 | 0 |
| Seq | 10 | 0 | 0 | 0 | 0 |
| Hill | 10 | 0 | 0 | 0 | 0 |
| Taboo | 10 | 0 | 0 | 0 | 0 |
| MB 14 | 4 | 0 | 2 | 0 | 4 |
| MB 16 | 6 | 0 | 1 | 0 | 3 |
| MB 18 | 2 | 0 | 0 | 1 | 7 |

**Pigs network results**

| Method | Best | > .9 | > .5 | > .01 | ≤ .01 |
|---|---|---|---|---|---|
| MPE | 0 | 0 | 0 | 0 | 10 |
| Hill | 0 | 0 | 0 | 0 | 10 |
| Taboo | 0 | 0 | 2 | 3 | 5 |
| ML | 5 | 1 | 3 | 1 | 0 |
| Hill | 9 | 0 | 1 | 0 | 0 |
| Taboo | 9 | 0 | 1 | 0 | 0 |
| Seq | 8 | 0 | 2 | 0 | 0 |
| Hill | 8 | 0 | 2 | 0 | 0 |
| Taboo | 8 | 0 | 2 | 0 | 0 |
| MB 14 | 3 | 0 | 3 | 4 | 0 |
| MB 16 | 1 | 1 | 4 | 3 | 1 |
| MB 18 | 6 | 0 | 2 | 2 | 0 |

**Water network results**

| Method | Best | > .9 | > .5 | > .01 | ≤ .01 |
|---|---|---|---|---|---|
| MPE | 9 | 0 | 1 | 0 | 0 |
| Hill | 10 | 0 | 0 | 0 | 0 |
| Taboo | 10 | 0 | 0 | 0 | 0 |
| ML | 8 | 1 | 1 | 0 | 0 |
| Hill | 10 | 0 | 0 | 0 | 0 |
| Taboo | 10 | 0 | 0 | 0 | 0 |
| Seq | 10 | 0 | 0 | 0 | 0 |
| Hill | 10 | 0 | 0 | 0 | 0 |
| Taboo | 10 | 0 | 0 | 0 | 0 |
| MB 14 | 6 | 1 | 2 | 0 | 1 |
| MB 16 | 6 | 1 | 2 | 1 | 0 |
| MB 18 | 9 | 0 | 0 | 1 | 0 |

Table 5: Detailed performance measures on the real world networks. Each column contains the number of times out of 10 that each algorithm was able to achieve the given performance relative to the best solution found.

## 4. Approximating MAP when Inference is Hard

The techniques developed thus far depend on the ability to perform exact inference. For many networks, even inference is intractable. In these cases, approximate inference can be substituted in order to produce MAP approximations.

We investigate using belief propagation as the approximate inference scheme, and local search for the optimization scheme. Iterative belief propagation is a useful approximate inference algorithm for approximating MAP for a number of reasons and has proven to be a very effective and efficient approximation method for a variety of domains. It has the ability to approximate MPE, posterior marginals, and probability of evidence, allowing for the same initialization schemes as we used for exact inference. Additionally, as we will show in section 4.2, after a single inference call, the scores of neighbors in the search space can be computed locally, allowing us to obtain the same linear speed up that we obtained using a similar approach in the exact inference case. Thus belief propagation allows all of the techniques for approximating MAP for inference tractable networks to be applied approximately when inference is not tractable.

### 4.1 Belief Propagation Review

Belief propagation was introduced as an exact inference method on polytrees (Pearl, 1988). It is a message passing algorithm in which each node in the network sends a message to its neighbors. These messages, along with the CPTs and the evidence can be used to compute posterior marginals for all of the variables. In networks with loops, belief propagation is no longer guaranteed to be exact, and successive iterations generally produce different results, so belief propagation is typically run until the message values converge. This has been shown to provide very good approximations for a variety of networks (McEliece, Rodemich, & Cheng, 1995; Murphy, Weiss, & Jordan, 1999), and has recently received a theoretical explanation (Yedidia, Freeman, & Weiss, 2000).

Belief propagation works as follows. Each node $X$, has an evidence indicator $\lambda_X$ where evidence can be entered. If the evidence sets $X = x$, then $\lambda_X(x) = 1$, and is 0 otherwise. If no evidence is set for $X$, then $\lambda_X(x) = 1$ for all $x$. After evidence is entered, each node $X$ sends a message to each of its neighbors. The message a node $X$ with parents $\mathbf{U}$ sends to child $Y$ is computed as

$$M_{XY} = \alpha \sum_{\mathbf{U}} \lambda_X \Pr(X|\mathbf{U}) \prod_{Z \neq Y} M_{ZX}$$

where $Z$ ranges over the neighbors of $X$ and $\alpha$ is a normalizing constant.[10] Similarly, the message $X$ sends to a parent $U$ is

$$M_{XU} = \alpha \sum_{X\mathbf{U}-\{U\}} \lambda_X \Pr(X|\mathbf{U}) \prod_{Z \neq U} M_{ZX}.$$

---

10. We use potential notation more common to join trees than the standard descriptions of belief propagation because we believe the many indices required in standard presentations mask the simplicity of the algorithm.

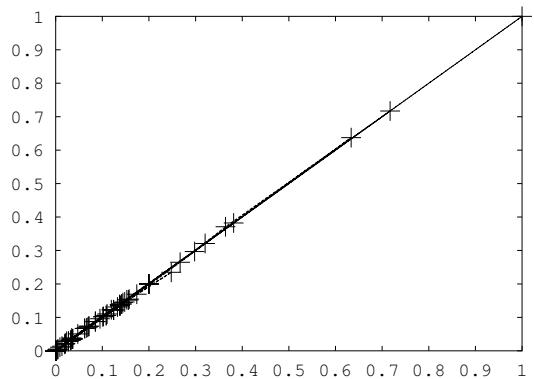

Figure 7: A scatter plot of the exact versus approximate retracted values of 30 variables of the Barley network. The x-axis is the true probability, and the y-axis the approximate probability.

Message passing continues until the message values converge. The posterior of $X$ is then approximated as

$$\Pr'(X|\mathbf{e}) = \alpha \sum_{\mathbf{U}} \lambda_X \Pr(X|\mathbf{U}) \prod_Z M_{ZX}.$$

The messages are all initialized to 1. There are two main schemes for ordering the messages. In the first scheme, all of the messages are computed simultaneously, based on the previous set of messages. In the other scheme, messages are updated incrementally, and in two phases, consistent with some ordering of the variables. In the first phase, in reverse order, each variable sends a message to its neighbors that precede it in the order. In the second phase, in order, each variable sends a message to its neighbors that come after it in the order. We implemented the second scheme since empirically it seems to converge faster than the first scheme (Murphy et al., 1999).

## 4.2 Approximating Neighbors' Scores

Belief propagation allows us to approximate the scores of neighbors in the local search space efficiently, similar to what we have done in the case of exact inference. The key as we shall show next is to be able to compute the quantity $\Pr(x|\mathbf{e} - X)$ for each variable $X$ efficiently, as we can use this quantity to rank the neighbors according to the desired score.

Specifically, in polytrees, the incoming messages are independent of the value of the local CPT or any evidence entered. Hence, leaving the evidence out of the product yields

$$\Pr(X|\mathbf{e} - X) = \alpha \sum_{\mathbf{U}} \Pr(X|\mathbf{U}) \prod_Z M_{ZX}.$$

Therefore, we can compute the above quantity for each variable after a single belief propagation. In networks that are not polytrees, the incoming messages are not necessarily independent of the evidence or the local CPT, but as is done with other BP methods, *we ignore that and hope that it is nearly independent.* Empirically, the approximation seems to be quite accurate. Figure 7 shows a representative example, comparing the correspondence between the approximate and exact retracted probabilities for 30 variables in the Barley network. The x axis corresponds to the true retracted probability, and the y axis to the approximation produced using belief propagation.

Still, $\Pr(x|\mathbf{e} - X)$ is not quite what we want to score neighbors in the local search space. But this quantity can be used to compute the ratio of the neighboring score to the current score which allows such comparisons. Specifically, simple algebra shows that:

$$\frac{\Pr(x, \mathbf{s} - X, \mathbf{e})}{\Pr(\mathbf{s}, \mathbf{e})} = \frac{\Pr(x|\mathbf{s} - X, \mathbf{e})}{\Pr(x_\mathbf{s}|\mathbf{s} - X, \mathbf{e})}$$

where $x_\mathbf{s}$ is the value that $X$ takes on in the current instantiation $\mathbf{s}$. Thus, we can find the neighbor with the best score after a single belief propagation.

## 4.3 Experimental Results

For the first experiment, we consider the improvement possible over what is typically done (MPE or ML) using it as a starting point and hill climbing from there. For the first experiment, we generated 100 synthetic networks with 100 variables each using the first method described in Appendix B with bias parameter 0.25 and width parameter of 13. We generated the networks to be small enough that we could often compute the exact MAP value, but large enough to make the problem challenging. We chose the MAP variables as the roots (typically between 20 and 25 variables), and the evidence values were chosen randomly from 10 of the leaves. We computed the true MAP for the ones which memory constraints (512 MB of RAM) allowed. We computed the true probability of the instantiations produced by the two standard methods. For both initialization methods we also computed the true probability of the instantiations returned by pure hill climbing[11] (i.e. only greedy steps were taken), and stochastic hill climbing with 100 steps, where random moves were taken with probability $p_f = .3$. Of the 100 networks, we were able to compute the exact MAP in 59 of them. Table 6 shows the number exactly solved for each method, as well as the worst instantiation produced, measured as the ratio of the probabilities of the found instantiation to the true MAP instantiation. All of the hill climbing methods improved significantly over their initializations in general, although for 2 of the networks, the hill climbing versions were slightly worse than the initial value (the worst was a ratio of .835), because of a slight mismatch in the true vs. approximate probabilities. Over all, the stochastic hill climbing routines outperformed the other methods.

In the second experiment, we generated 25 random MAP problems for the Barley network, each with 25 randomly chosen MAP variables, and 10 randomly chosen evidence assignments. We use the same parameters as in the previous experiment. The problems were too hard to compute the exact MAP, so we report only on the relative improvements over the initialization methods. Table 7 summarizes the results. Again, the stochastic hill climbing methods were able to significantly improve the quality of the instantiations created.

---

11. We compare pure and stochastic hill climbing to evaluate what can be gained by stochastic methods. The initial hill climb usually requires very few evaluations, so if stochastic methods make little difference, efficiency considerations would dictate that pure hill climbing be used.

|           | # solved exactly | worst |
|-----------|------------------|-------|
| MPE       | 9                | .015  |
| MPE-Hill  | 41               | .06   |
| MPE-SHill | 43               | .21   |
| ML        | 31               | .34   |
| ML-Hill   | 38               | .46   |
| ML-SHill  | 42               | .72   |

Table 6: Solution quality for the random networks. Shows the number solved exactly of the 59 for which we could compute the true MAP value. Worst is the ratio of the probabilities of the found instantiation to the true MAP instantiation. Each hill climbing method improved significantly over the initializations.

|           | min        | median      | mean             | max              |
|-----------|------------|-------------|------------------|------------------|
| MPE-Hill  | 1.0        | 8.4         | $1.3\times10^{11}$ | $3.1\times10^{12}$ |
| MPE-SHill | 1.0        | 8.4         | $1.3\times10^{11}$ | $3.1\times10^{12}$ |
| ML-Hill   | $1.0\times10^{4}$ | $3.6\times10^{7}$ | $3.4\times10^{15}$ | $8.4\times10^{16}$ |
| ML-SHill  | $7.7\times10^{3}$ | $3.6\times10^{7}$ | $3.4\times10^{15}$ | $8.4\times10^{16}$ |

Table 7: The statistics on the improvement over just the initialization method for each search method on the data set generated from the Barley network. Improvement is measured as the ratio of the found probability to the probability of the initialization instantiation.

In the third experiment, we performed the same type of experiment on the Pigs network. None of the search methods were able to improve on ML initialization. We concluded that the problem was too easy. Pigs has over 400 variables, and it seemed that the evidence didn't force enough dependence among the variables. We ran another experiment with Pigs, this time using 200 MAP variables and 20 evidence values to make it more difficult. Table 8 summarizes the results. Again, the stochastic methods were able to improve significantly over the initialization methods.

|           | min  | median      | mean             | max              |
|-----------|------|-------------|------------------|------------------|
| MPE-Hill  | 1.0  | $1.7\times10^{5}$ | $1.5\times10^{7}$ | $3.3\times10^{8}$ |
| MPE-SHill | 1.0  | $2.5\times10^{5}$ | $4.5\times10^{11}$ | $1.1\times10^{13}$ |
| ML-Hill   | 13.0 | $2.0\times10^{3}$ | $3.3\times10^{5}$ | $4.5\times10^{6}$ |
| ML-SHill  | 13.0 | $1.2\times10^{4}$ | $8.2\times10^{5}$ | $8.2\times10^{6}$ |

Table 8: The statistics on the improvement over just the initialization method alone for each search method on the data set generated from the Pigs network. Improvement is measured as the ratio of the found probability to the initialization probability.

|  | MPE | | | ML | | | Seq | | | MB | | |
|---|---|---|---|---|---|---|---|---|---|---|---|---|
|  | I | H | T | I | H | T | I | H | T | 14 | 16 | 18 |
| Barley | 2 | 6 | 7 | 2 | 7 | 7 | 7 | 9 | 9 | 1 | 3 | 5 |
| Mildew | 5 | 1 | 10 | 5 | 0 | 10 | 9 | 0 | 10 | 4 | 4 | 7 |
| Munin2 | 5 | 0 | 9 | 5 | 0 | 9 | 10 | 0 | 10 | 4 | 5 | 7 |
| Munin3 | 8 | 0 | 1 | 8 | 0 | 1 | 10 | 0 | 1 | 4 | 6 | 2 |
| Pigs | 0 | 0 | 1 | 0 | 0 | 1 | 8 | 0 | 8 | 3 | 1 | 6 |
| Water | 9 | 0 | 0 | 9 | 0 | 0 | 10 | 0 | 0 | 6 | 6 | 9 |

Table 9: Number of times out of ten that each algorithm found the instantiation that yielded the highest score. The I, H and T entries stand for initial, hill climbing, and taboo respectively. MB stands for mini–buckets, and 14, 16, and 18 are the width bounds.

We also ran these algorithms on the same queries on the real world networks that were used in Section 3.4 to be able to compare performance between the methods. Table 9 shows how they performed and compares their performance to the mini–bucket algorithms. Table 10 gives a more detailed exposition of their performance. There are a couple of interesting items about this data set. One is the surprising performance of simple sequential initialization. Over all, it performed the best of the approximate algorithms. Another interesting thing to note is that hill climbing often negatively impacted performance. This suggests that marginal computations are often more accurate than probability of evidence computations. This problem is especially acute in networks with significant determinism. While belief propagation believes a configuration has significant probability, it may actually have 0 probability because one of its constraints is violated. These experiments suggest that it is possible to improve on the standard approaches used when inference is intractable (approximating MPE, or ML or using a mini–bucket scheme) by using belief propagation to estimate the joint, and successively moving to states with higher approximate scores.

## 5. Conclusion

MAP is a computationally very hard problem which is not in general amenable to exact solution even for very restricted classes (ex. polytrees). Even approximation is difficult. Still, we can produce approximations that are much better than those currently used by practitioners (MPE, ML) through using approximate optimization and inference methods. We showed one method based on belief propagation and stochastic hill climbing that produced significant improvements over those methods, extending the realm for which MAP can be approximated to networks that work well with belief propagation.

## Acknowledgement

This work has been partially supported by MURI grant N00014-00-1-0617

**Barley network results**

| Method | Best | > .9 | > .5 | > .01 | ≤ .01 |
|--------|------|------|------|-------|-------|
| MPE | 2 | 3 | 1 | 3 | 1 |
| Hill | 6 | 1 | 1 | 2 | 0 |
| Taboo | 7 | 0 | 1 | 2 | 0 |
| ML | 2 | 3 | 1 | 3 | 1 |
| Hill | 7 | 1 | 1 | 1 | 0 |
| Taboo | 7 | 0 | 1 | 2 | 0 |
| SEQ | 7 | 1 | 1 | 1 | 0 |
| Hill | 9 | 0 | 1 | 0 | 0 |
| Taboo | 9 | 0 | 1 | 0 | 0 |
| MB 14 | 1 | 2 | 1 | 3 | 3 |
| MB 16 | 3 | 3 | 1 | 2 | 1 |
| MB 18 | 5 | 2 | 0 | 3 | 0 |

**Mildew network results**

| Method | Best | > .9 | > .5 | > .01 | ≤ .01 |
|--------|------|------|------|-------|-------|
| MPE | 5 | 2 | 3 | 0 | 0 |
| Hill | 1 | 0 | 0 | 0 | 9 |
| Taboo | 10 | 0 | 0 | 0 | 0 |
| ML | 5 | 2 | 3 | 0 | 0 |
| Hill | 0 | 0 | 0 | 0 | 10 |
| Taboo | 10 | 0 | 0 | 0 | 0 |
| Seq | 9 | 1 | 0 | 0 | 0 |
| Hill | 0 | 0 | 0 | 0 | 10 |
| Taboo | 10 | 0 | 0 | 0 | 0 |
| MB 14 | 4 | 1 | 0 | 1 | 4 |
| MB 16 | 4 | 0 | 0 | 1 | 5 |
| MB 18 | 7 | 0 | 1 | 0 | 2 |

**Munin2 network results**

| Method | Best | > .9 | > .5 | > .01 | ≤ .01 |
|--------|------|------|------|-------|-------|
| MPE | 5 | 0 | 3 | 1 | 1 |
| Hill | 0 | 0 | 0 | 0 | 10 |
| Taboo | 9 | 0 | 0 | 0 | 1 |
| ML | 5 | 0 | 3 | 1 | 1 |
| Hill | 0 | 0 | 0 | 0 | 10 |
| Taboo | 9 | 0 | 0 | 0 | 1 |
| Seq | 10 | 0 | 0 | 0 | 0 |
| Hill | 0 | 0 | 0 | 0 | 10 |
| Taboo | 10 | 0 | 0 | 0 | 0 |
| MB 14 | 4 | 0 | 1 | 2 | 3 |
| MB 16 | 5 | 0 | 1 | 2 | 2 |
| MB 18 | 7 | 0 | 0 | 1 | 2 |

**Munin3 network results**

| Method | Best | > .9 | > .5 | > .01 | ≤ .01 |
|--------|------|------|------|-------|-------|
| MPE | 8 | 0 | 0 | 1 | 1 |
| Hill | 0 | 0 | 0 | 0 | 10 |
| Taboo | 1 | 0 | 0 | 0 | 9 |
| ML | 8 | 0 | 0 | 1 | 1 |
| Hill | 0 | 0 | 0 | 0 | 10 |
| Taboo | 1 | 0 | 0 | 0 | 9 |
| Seq | 10 | 0 | 0 | 0 | 0 |
| Hill | 0 | 0 | 0 | 0 | 10 |
| Taboo | 1 | 0 | 0 | 0 | 9 |
| MB 14 | 4 | 0 | 2 | 0 | 4 |
| MB 16 | 6 | 0 | 1 | 0 | 3 |
| MB 18 | 2 | 0 | 1 | 0 | 7 |

**Pigs network results**

| Method | Best | > .9 | > .5 | > .01 | ≤ .01 |
|--------|------|------|------|-------|-------|
| MPE | 0 | 0 | 0 | 0 | 10 |
| Hill | 0 | 0 | 0 | 0 | 10 |
| Taboo | 1 | 0 | 0 | 4 | 5 |
| ML | 0 | 0 | 0 | 0 | 10 |
| Hill | 0 | 0 | 0 | 0 | 10 |
| Taboo | 1 | 0 | 0 | 4 | 5 |
| Seq | 8 | 0 | 2 | 0 | 0 |
| Hill | 0 | 0 | 0 | 0 | 10 |
| Taboo | 8 | 0 | 2 | 0 | 0 |
| MB 14 | 3 | 0 | 3 | 4 | 0 |
| MB 16 | 1 | 1 | 4 | 3 | 1 |
| MB 18 | 6 | 0 | 2 | 2 | 0 |

**Water network results**

| Method | Best | > .9 | > .5 | > .01 | ≤ .01 |
|--------|------|------|------|-------|-------|
| MPE | 9 | 0 | 1 | 0 | 0 |
| Hill | 0 | 0 | 0 | 0 | 10 |
| Taboo | 0 | 0 | 0 | 0 | 10 |
| ML | 9 | 0 | 1 | 0 | 0 |
| Hill | 0 | 0 | 0 | 0 | 10 |
| Taboo | 0 | 0 | 0 | 0 | 10 |
| Seq | 10 | 0 | 0 | 0 | 0 |
| Hill | 0 | 0 | 0 | 0 | 10 |
| Taboo | 0 | 0 | 0 | 0 | 10 |
| MB 14 | 6 | 1 | 2 | 0 | 1 |
| MB 16 | 6 | 1 | 2 | 1 | 0 |
| MB 18 | 9 | 0 | 0 | 1 | 0 |

Table 10: Detailed performance measures on the real world networks using Belief propagation approximation methods. Each column contains the number of times out of 10 that each algorithm was able to achieve the given performance relative to the best solution found.

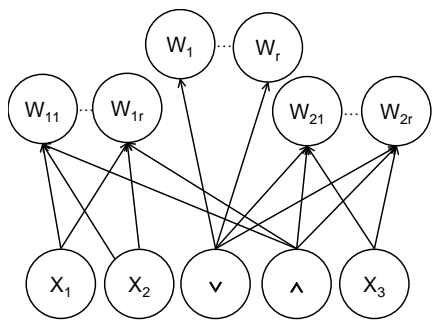

Figure 8: The network produced using the construction in the proof of Theorem 2 for the formula $(x_1 \wedge x_2) \vee x_3$.

# Appendix A. Proofs of Theorems

## Proof of Theorem 2

We want to show that MAP remains $NP^{PP}$-complete even when restricted to networks of depth 2, with no evidence, only binary variables, and parameters that are arbitrarily close to $1/2$. Membership in $NP^{PP}$ was established in Theorem 1. We show hardness by providing a reduction from **E-MAJSAT**.

The flow of the proof is as follows. First, we construct a depth 2 Bayesian network from the **E-MAJSAT** problem. Then, we show that by asserting some evidence, we can overcome the constraint that all of the parameters lie within $[1/2 - \epsilon, 1/2 + \epsilon]$, and use MAP to obtain the **E-MAJSAT** solution. Finally, we show that by including the evidence variables as MAP variables instead, no evidence is needed.

The network is constructed as follows. Each logical variable $x_i$ induces a network variable $X_i$ with uniform prior. Each operand $y_i$ induces a network variable $Y_i$ with a uniform prior. Notice that they are not connected, so unlike the reduction in Theorem 1, the CPT entries do not enforce that an operator variable take on a value that is consistent its operands with respect to the to the logic of the formula. For example, the network will assign positive probability to an "and" node being true, and both of its operand variables being false. We say that a variable $Y_i$ is consistent with the variables $P_i$ associated with its operands, if the logical function of operator $y_i$ yields the value of $Y_i$ on input $\mathbf{p}_i$. Consistency, instead of being enforced rigidly, is weighted by introducing $r$ weight variables $W_{i1}...W_{ir}$ (the actual value of $r$ will be discussed subsequently) associated with each $Y_i$. The parents of $W_{ij}$ are the operator variable $Y_i$ and the variables corresponding to its operands. The CPT of $W_{ij}$ is defined as

$$\Pr(W_{ij} = T | Y_i, P_i) = \begin{cases} \frac{1}{2} + \epsilon & Y_i \text{ is consistent with } P_i \\ \frac{1}{2} & \text{otherwise} \end{cases}$$

where $P_i$ are the variables associated with the operands of $y_i$. Finally, as children of $Y_m$ (which corresponds to the top level operator) we add $r$ additional binary variables $W_1...W_r$,

where

$$\Pr(W_j = T | Y_m) = \begin{cases} \frac{1}{2} + \epsilon & \text{if } Y_m = T \\ \frac{1}{2} & \text{otherwise} \end{cases}$$

for the purpose of weighting states in which the formula is satisfied. See Figure 8 for an example network construction.

Now, consider the probability of a complete instantiation of the variables, where all of the weight variables (which includes both the consistency weighting variables $W_{ij}$, and satisfiability weighting variables $W_i$) are set to true, which we denote as $\mathbf{W} = \mathbf{T}$.

$$\Pr(\mathbf{x}, \mathbf{y}, \mathbf{W} = \mathbf{T}) = \left(\frac{1}{2}\right)^{m+n} \left(\frac{1}{2} + \epsilon\right)^{kr} \left(\frac{1}{2}\right)^{(m-k)r} \left(\frac{1}{2} + \epsilon\right)^{sr} \left(\frac{1}{2}\right)^{(1-s)r}$$

where $\mathbf{x}$ is an instantiation of $X_1...X_n$, $\mathbf{y}$ is an instantiation of $\mathbf{Y}_1...\mathbf{Y}_m$, $k$ is the number of operator variables that are consistent with their operands variables and s=1 if $Y_m = T$, 0 otherwise. For a consistent satisfying assignment $\mathbf{xy}$,

$$\Pr(\mathbf{x}, \mathbf{y}, \mathbf{W} = \mathbf{T}) = \left(\frac{1}{2}\right)^{m+n} \left(\frac{1}{2} + \epsilon\right)^{(m+1)r}$$

while for an inconsistent, or unsatisfying assignment $\mathbf{xy}$,

$$\Pr(\mathbf{x}, \mathbf{y}, \mathbf{W} = \mathbf{T}) \leq \left(\frac{1}{2}\right)^{m+n+r} (\frac{1}{2} + \epsilon)^{mr}.$$

We want to choose $r$ such that the probability of a single consistent satisfying instance is greater than twice the sum of all of the probabilities of inconsistent or unsatisfying instances. The number of inconsistent or unsatisfying instances is bounded by $2^{n+m}$, so we want an $r$ where

$$\left(\frac{1}{2}\right)^{m+n} \left(\frac{1}{2} + \epsilon\right)^{(m+1)r} > 2^{n+m+1} \left(\frac{1}{2}\right)^{m+n+r} (\frac{1}{2} + \epsilon)^{mr}.$$

Solving for $r$ yields

$$r > \frac{(m+n+1)}{1 + \log_2\left(\frac{1}{2} + \epsilon\right)}$$

which is linear in the size of the formula, so the size of the reduction remains polynomially bounded.

Let $C = \Pr(\mathbf{x}, \mathbf{y}, \mathbf{W} = \mathbf{T}) = \left(\frac{1}{2}\right)^{m+n} \left(\frac{1}{2} + \epsilon\right)^{(m+1)r}$ where $\mathbf{xy}$ is a consistent satisfying instance. Then, for a particular instantiation $\mathbf{q}$ of $X_1...X_k$,

$$\begin{aligned} \Pr(\mathbf{q}, \mathbf{W} = \mathbf{T}) &= \sum_{x_{k+1},...,x_n, y_1,...,y_m} \Pr(\mathbf{q}, x_{k+1},...x_n, y_1,...y_m, \mathbf{W} = \mathbf{T}) \\ &= \#_\mathbf{q} C + \sum_{\mathbf{xy}} \Pr(\mathbf{xy}, W = \mathbf{T}) \end{aligned}$$

where $\#_\mathbf{q}$ is the number of complete variable instantiations compatible with $\mathbf{q}$ that satisfies $\phi$ and $\mathbf{xy}$ ranges over the inconsistent and unsatisfying assignments compatible with $x_1...x_k$. Since for any instantiation of $\mathbf{x}$ there is only one compatible instantiation, $\#_\mathbf{q}$ also corresponds to the number of satisfiable instantiations of $\phi$ consistent with $\mathbf{q}$. The choice of $r$ ensures that the sum of the unsatisfying or inconsistent instantiations is less than $C/2$, and is always greater than 0 assuming there is at least one operator (since there is some instantiation where the operator and its operands are not consistent). Thus $\#_q C < \Pr(\mathbf{q}, \mathbf{W} = \mathbf{T}) < (\#_q + 1/2)C$. There are $2^{n-k}$ possible instantiations of $X_{k+1}...X_n$, so if half or less are satisfied then $\Pr(\mathbf{q}, \mathbf{W} = T) < (2^{n-k-1} + 1/2)C$, while if more than half are satisfied then $\Pr(\mathbf{q}, \mathbf{W} = T) > (2^{n-k-1} + 1)C$. Thus the **D-MAP** query, using MAP variables $X_1...X_k$, evidence $\mathbf{W} = \mathbf{T}$, and threshold $(2^{n-k-1} + 1)C$ is true if and only if the **E-MAJSAT** query is also true.

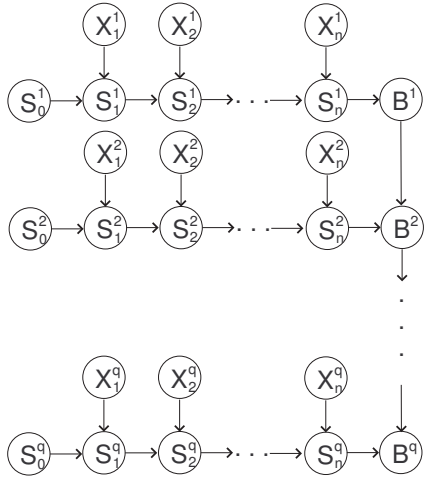

Figure 9: The network used in the reduction of Theorem 8.

Now, notice that in every table that contains a weight variable, the value of the configuration where it takes on true is greater or equal to the value when it takes on false. Thus $\Pr(\mathbf{q}, \mathbf{W} = \mathbf{T}) \geq \Pr(\mathbf{q}, \mathbf{W} = \mathbf{w})$, for all $\mathbf{q}$ and $\mathbf{w}$. It then follows that $\mathrm{MAP}(X_1...X_k, \mathbf{W} = \mathbf{T}) = \mathrm{MAP}(X_1...X_k\mathbf{W}, \emptyset)$. Therefore, the **D-MAP** query, using MAP variables $X_1...X_k, \mathbf{W}$ and no evidence, with threshold $(2^{n-k-1} + 1)C$ is true if and only if the **E-MAJSAT** query is also true. $\square$

**Proof of Theorem 8**

As part of the proof of the theorem, we will use the following lemma.

**Lemma 9** *For all $x \geq 1$, $4x + \frac{1}{2} > \frac{1}{\ln(1+\frac{1}{4x})}$.*

**Proof:** First, we show that $f(x) = \ln(1 + \frac{1}{4x}) - \frac{1}{4x+\frac{1}{2}}$ is monotonically decreasing for $x \geq 1$.

$$\frac{df}{dx} = \frac{-1}{4x^2 + x} + \frac{4}{(4x + \frac{1}{2})^2}$$

$$= \frac{16x^2 + 4x - (4x + \frac{1}{2})^2}{(4x + \frac{1}{2})^2(4x^2 + x)}$$

$$= \frac{-1}{4(4x + \frac{1}{2})^2(4x^2 + x)}$$

which is always negative for $x \geq 1$, and hence $f(x)$ is monotonically decreasing.

Now, since $f(x)$ is monotonically decreasing, and $\lim_{x \to \infty} f(x) = 0$, $f(x)$ must be strictly positive. Thus, for all $x \geq 1$, $\ln(1 + \frac{1}{4x}) > \frac{1}{4x + \frac{1}{2}}$ which implies $4x + \frac{1}{2} > \frac{1}{\ln(1 + \frac{1}{4x})}$. $\quad \square$

The basic idea of the proof is to show by repeating the construction of Theorem 7 a polynomial number of times, that if we can approximate MAP on polytrees within relative error $2^{\text{size}^\epsilon}$ for any $\epsilon \in [0, 1)$, where the size of the network is parameterized by the number of conditional probability parameters, then we can solve SAT in polynomial time.

Given a SAT problem instance with $n$ variables, and $m$ clauses, we create a Bayesian network by replicating the construction from Theorem 7 q times, and connecting them to form a polytree. Specifically, to each copy $i$ of the construction, we add a variable $B^i$ (we use superscripts to denote variables associated with a particular copy of the construction), with parents $S_n^i$, and if $i > 1$, parent $B^{i-1}$ (see Figure 9). The conditional probability of $B^i$ is uniform for all parent instantiations. We choose $q$ to satisfy $(1 + \frac{1}{4m})^q > 2^{\text{size}^\epsilon}$. We now show that $q$ can be chosen so that the network size remains polynomial in the size of the logical formula. The resulting network has $q(2n + 2)$ variables, and each conditional probability table has at most $2(m + 1)^2$ parameters, so the total size of the reduction is bounded by $q(m + 1)^2(4n + 4)$. Replacing size with the size bound places the constraint

$$\left(1 + \frac{1}{4m}\right)^q > 2^{\left(q(m+1)^2(4n+4)\right)^\epsilon}$$

on $q$. Since $0 \leq \epsilon < 1$, solving for $q$ yields

$$\left(1 + \frac{1}{4m}\right)^q > 2^{\left(q(m+1)^2(4n+4)\right)^\epsilon}$$

$$q \ln\left(1 + \frac{1}{4m}\right) > q^\epsilon (m+1)^{2\epsilon}(4n+4)^\epsilon \ln 2$$

$$q^{1-\epsilon} > \frac{(m+1)^{2\epsilon}(4n+4)^\epsilon \ln 2}{\ln\left(1 + \frac{1}{4m}\right)}$$

$$q > \left(\frac{(m+1)^{2\epsilon}(4n+4)^\epsilon \ln 2}{\ln\left(1 + \frac{1}{4m}\right)}\right)^{\frac{1}{1-\epsilon}}$$

Now, from Lemma 9, $4m + 1/2 > 1/\ln(1 + 1/4m)$, so substitution yields a stronger bound,

$$q > \left(\left(4m + \frac{1}{4}\right)(m+1)^{2\epsilon}(4n+4)^\epsilon \ln 2\right)^{\frac{1}{1-\epsilon}}$$

which is polynomially bounded. Thus the network can be constructed in time polynomial in the size of the formula.

Then, for a particular instantiation $\mathbf{x}$ of all $X$ variables $X_1^1 ... X_n^q$, and evidence $\mathbf{s}$ asserting $S_n^i = 0$ for each $i$, the probability is

$$\Pr(\mathbf{x}, \mathbf{s}) = \prod_i \Pr(\mathbf{x}^i, S_n^i = 0)$$

$$= \prod_i \frac{\# \text{ clauses satisfied by } \mathbf{x}^i}{m2^n}$$

because each subnetwork is independent. Thus the solution $M$ to MAP over $X_1^1 ... X_n^q$ with evidence $S_n^1 = .... = S_n^q = 0$ is $\left(\frac{k}{m2^n}\right)^q$ where $k$ is the maximum number of clauses that can be simultaneously satisfied in the original SAT problem. If the problem is satisfiable then $k = m$, and so the approximate solution $M'$ obeys

$$M' \geq \frac{M}{2^{\text{size}^\epsilon}} > \left(\frac{4m}{4m+1}\right)^q \left(\frac{m}{m2^n}\right)^q > \left(\frac{m - \frac{1}{4}}{m}\right)^q \left(\frac{1}{2^n}\right)^q = \left(\frac{m - \frac{1}{4}}{m2^n}\right)^q$$

On the other hand, if it isn't satisfiable, then $k \leq m - 1$, so

$$M' \leq 2^{\text{size}^\epsilon} M < \left(\frac{4m+1}{4m}\right)^q \left(\frac{m-1}{m2^n}\right)^q = \left(\frac{(4m+1)(m-1)}{4m}\right)^q \left(\frac{1}{m2^n}\right)^q < \left(\frac{m - \frac{3}{4}}{m2^n}\right)^q$$

The upper bound of $M'$ if the SAT problem is unsatisfiable is bounded below the lower bound of $M'$ if it is satisfiable. Because the network construction and the bound tests can be accomplished in polynomial time, if the MAP problem itself can be approximated within a factor of $2^{\text{size}^\epsilon}$ in polynomial time then SAT can be decided in polynomial time.

## Appendix B. Generating Random Networks

We generated several types of networks to perform our experiments. We used two methods for generating the structure, and a single parametric method for generating the quantification.

### B.1 Generating the Network Structure

The first method is parameterized by the number of variables $N$ and the connectivity $c$. This method tends to produce structures with widths that are close to $c$. Darwiche (2001) provides an algorithmic description.

The second method is parameterized by the number of variables $N$, and the probability $p$ of an edge being present. We generate an ordered list of $N$ variables, and add an edge between variables $X$ and $Y$ with probability $p$. The edges added are directed toward the variable that appears later in the order.

### B.2 Quantifying the Dependencies

The quantification method is parameterized by a bias parameter $b$. The values of the CPTs for the roots were chosen uniformly. The values for the rest of the nodes were based on a bias, where one of the values $v$ was chosen uniformly in $[0, b)$, and the other as $1 - v$. For example, for $b = .1$, each non root variable given its parents has one value in $[0, .1)$, and the other in $(.9, 1]$. Special cases $b = 0$, and $b = .5$ produce deterministic, and uniformly random quantifications respectively.

# References

Dagum, P., & Luby, M. (1997). An optimal approximation algorithm for Bayesian inference. *Artificial Intelligence*, *93*, 1–27.

Darwiche, A. (2001). Recursive conditioning. *Artificial Intelligence*, *126*(1-2), 5–41.

Darwiche, A. (2003). A differential approach to inference in Bayesian networks. *Journal of the ACM*, *50*(3), 280–305.

de Campos, L., Gamez, J., & Moral, S. (1999). Partial abductive inference in Bayesian belief networks using a genetic algorithm. *Pattern Recognition Letters*, *20(11-13)*, 1211–1217.

Dechter, R., & Rish, I. (1998). Mini-buckets: A general scheme for approximate inference. Tech. rep. R62a, Information and Computer Science Department, UC Irvine.

Dechter, R. (1996). Bucket elimination: A unifying framework for probabilistic inference. In *Proceedings of the 12th Conference on Uncertainty in Artificial Intelligence (UAI)*, pp. 211–219.

Huang, C., & Darwiche, A. (1996). Inference in belief networks: A procedural guide. *International Journal of Approximate Reasoning*, *15*(3), 225–263.

Jensen, F. V., Lauritzen, S., & Olesen, K. (1990). Bayesian updating in recursive graphical models by local computation. *Computational Statistics Quarterly*, *4*, 269–282.

Kask, K., & Dechter, R. (1999). Stochastic local search for Bayesian networks. In *Seventh International Workshop on Artificial Intelligence*, Fort Lauderdale, FL. Morgan Kaufmaann.

Kjaerulff, U. (1990). Triangulation of graphs—algorithms giving small total state space. Tech. rep. R-90-09, Department of Mathematics and Computer Science, University of Aalborg, Denmark.

Lauritzen, S. L., & Spiegelhalter, D. J. (1988). Local computations with probabilities on graphical structures and their application to expert systems. *Journal of Royal Statistics Society, Series B*, *50*(2), 157–224.

Litmman, M., Majercik, S. M., & Pitassi, T. (2001). Stochastic boolean satisfiability. *Journal of Automated Reasoning*, *27*(3), 251–296.

Littman, M. (1999). Initial experiments in stochastic satisfiability. In *Sixteenth National Conference on Artificial Intelligence*, pp. 667–672.

Littman, M., Goldsmith, J., & Mundhenk, M. (1998). The computational complexity of probabilistic planning.. *Journal of Artificial Intelligence Research*, *9*, 1–36.

McEliece, R. J., Rodemich, E., & Cheng, J. F. (1995). The turbo decision algorithm. In *33rd Allerton Conference on Communications, Control and Computing*, pp. 366–379.

Mengshoel, O. J., Roth, D., & Wilkins, D. C. (2000). Stochastic greedy search: Efficiently computing a most probable explanation in Bayesian networks. Tech. rep. UIUCDS-R-2000-2150, U of Illinois Urbana-Champaign.

Murphy, K. P., Weiss, Y., & Jordan, M. I. (1999). Loopy belief propagation for approximate inference: an emperical study. In *Proceedings of Uncertainty in AI*.

Papadimitriou, C., & Tsitsiklis, J. (1987). The complexity of Markov decision processes. *Mathematics of Operations Research*, *12(3)*, 441–450.

Park, J., & Darwiche, A. (2003). A differential semantics for jointree algorithms. In *Neural Information Processing Systems (NIPS) 15*.

Pearl, J. (1988). *Probabilistic Reasoning in Intelligent Systems: Networks of Plausible Inference*. Morgan Kaufmann Publishers, Inc., San Mateo, California.

Roth, D. (1996). On the hardness of approximate reasoning. *Artificial Intelligence*, *82*(1-2), 273–302.

Shenoy, P. P., & Shafer, G. (1986). Propagating belief functions with local computations. *IEEE Expert*, *1*(3), 43–52.

Shimony, S. E. (1994). Finding MAPs for belief networks is NP–hard. *Artificial Intelligence*, *68*(2), 399–410.

Toda, S. (1991). PP is as hard as the polynomial-time hierarchy. *SIAM Journal of Computing*, *20*, 865–877.

Yedidia, J., Freeman, W., & Weiss, Y. (2000). Generalized belief propagation. In *NIPS*, Vol. 13.